\documentclass[10pt,twocolumn,letterpaper]{article}

\usepackage{cvpr}
\usepackage{times}
\usepackage{epsfig}
\usepackage{graphicx}
\usepackage{amsmath}
\usepackage{amssymb}
\usepackage{multirow}
\usepackage{subcaption}
\usepackage[pagebackref=true,breaklinks=true,letterpaper=true,colorlinks,bookmarks=false]{hyperref}

\cvprfinalcopy % *** Uncomment this line for the final submission

\def\cvprPaperID{****} % *** Enter the CVPR Paper ID here

\ifcvprfinal\fi
\begin{document}

%%%%%%%%% TITLE
\title{ShopSign: a Diverse Scene Text Dataset of Chinese Shop Signs in Street Views}

\author{Chongsheng Zhang\\
Henan University\\
475001 KaiFeng, China\\
{\tt\small chongsheng.zhang@yahoo.com}
\and
Guowen Peng\\
Henan University\\
475001 KaiFeng, China\\
{\tt\small gwpeng@henu.edu.cn}
\and
Yuefeng Tao\\
Henan University\\
475001 KaiFeng, China\\
{\tt\small yftao@henu.edu.cn}
\and
Feifei Fu\\
Henan University\\
475001 KaiFeng, China\\
{\tt\small fff@henu.edu.cn}
\and
Wei Jiang\\
North China University of Water\\
Resources and Electric Power\\
450045 Zhengzhou, China\\
{\tt\small jiangwei@ncwu.edu.cn}
\and
George Almpanidis\\
Henan University\\
475001 KaiFeng, China\\
{\tt\small almpanidis@gmail.com}
\and
Ke Chen\\
South China University of Technology\\
510006 Guangzhou, China\\
{\tt\small chenk@scut.edu.cn}
}
\maketitle
%\thispagestyle{empty}

%%%%%%%%% ABSTRACT
\begin{abstract}
In this paper, we introduce the ShopSign dataset, which is a newly developed natural scene text dataset of Chinese shop signs in street views. Although a few scene text datasets are already publicly available (e.g. ICDAR2015, COCO-Text), there are few images in these datasets that contain Chinese texts/characters. Hence, we collect and annotate the ShopSign dataset to advance research in Chinese scene text detection and recognition. 

The new dataset has three distinctive characteristics: (1) large-scale: it contains 25,362 Chinese shop sign images, with a total number of 196,010 text-lines. (2) diversity: the images in ShopSign were captured in different scenes, from downtown to developing regions, using more than 50 different mobile phones. (3) difficulty: the dataset is very sparse and imbalanced. It also includes five categories of hard images (mirror, wooden, deformed, exposed and obscure). To illustrate the challenges in ShopSign, we run baseline experiments using state-of-the-art scene text detection methods (including CTPN, TextBoxes++ and EAST), and cross-dataset validation to compare their corresponding performance on the related datasets such as CTW, RCTW and ICPR 2018 MTWI challenge dataset.

The sample images and detailed descriptions of our ShopSign dataset are publicly available at: \url{https://github.com/chongshengzhang/shopsign}.

\end{abstract}

%%%%%%%%% BODY TEXT
\section{Introduction}

Datasets, deep learning (DL) techniques and high-performance computing facilities are the three most important driving forces in today's new AI era.  DL has also outperformed the conventional methods on scene text detection, which  has received significant attention in the past decade. However, existing solutions all focus on English (Latin) language, and little effort has been invested in Chinese scene text detection and recognition (hereafter referred to as ``Chinese Photo OCR"). This is not only because of the difficulty of Chinese text recognition owing to the complexity of the language and its huge number of classes / characters (the number of commonly used Chinese characters is 6,763, while English language only has 26 letters), but also due to the scarcity of large-scale well-annotated datasets of Chinese natural scene images, since deep learning based techniques are data-driven, data-eager. 

 In the literature, early approaches to scene text detection use ``low-level" features to localize texts in natural scene images.  Starting from 2012, growing effort has been devoted to the development of ``high-level" deep learning based scene text detection approaches, which have shown significantly better performance than conventional methods.  CTPN~\cite{CTPN} and TextBoxes~\cite{Textboxes} are representative methods for horizontal text detection, while EAST~\cite{EAST} and TextBoxes++~\cite{TextBoxesplus} are recent solutions for multi-oriented text detection. However, most of these advances are towards scene text detection, and only until recent years, researchers have started investigating DL-based scene text recognition, where CRNN~\cite{CRNN} and Sliding CNN~\cite{SlidingCNN} are representative solutions. Nevertheless,  few methods have been designed towards  Chinese scene text detection and recognition. 

To advance research in Chinese Photo OCR, we present a diverse and challenging dataset of Chinese natural scene images, which consists of shop signs along the streets in China. This dataset is hereafter referred to as ``ShopSign". It contains  25,770 images collected from more than 20 cities, using 50 different smart phones. These images exhibit a wide variety of scales, orientations, lighting conditions, layout and geo-spatial locations as well as class imbalance. Moreover, we  characterize the difficulty of ShopSign by specifying 5 categories of ``hard" images, which contain mirror, exposed, obscured, wooden, or deformed texts. The images in ShopSign have been manually annotated in ``text-line-based" manner by 10 research assistants. 

Upon ShopSign, we train and evaluate well-known deep network architectures (as baseline models) and report their text detection/recognition performance. We show and analyze the deficiency of existing deep-learning based Photo OCR solutions on ShopSign, and point out that Chinese OCR is a very difficult task and needs significantly more research effort. We hope the publicity of this large ShopSign dataset will spur new advances in Chinese scene text detection and recognition.

The main contributions of this work are summarized as follows:

\begin{enumerate}
\item
We present ShopSign, which is a large-scale, diverse and difficult dataset of Chinese scene images for  text detection and recognition.
\item
We report results of several baseline methods on ShopSign, including EAST, TextBoxes++, and CTPN. Through cross-dataset experiments, we observe  improved results on difficult detection/recognition cases with ShopSign.
\end{enumerate}

\section{Related Work}

In this section, we introduce state-of-the-art algorithms in scene text detection and recognition. We will also present related  datasets.

\subsection{Scene Text Detection and Recognition}

A scene text recognition system usually consists of two main components: scene text detector and recognizer. The former module localizes characters/texts in images in the form of bounding boxes, while the latter identifies texts (character sequences) from the cropped images inside the bounding boxes. But there are also a few attempts that aim to directly output the text in an ``end-to-end" manner, i.e., seamlessly  integrating  scene text detection and recognition in a single neural network (procedure). However, we argue that it is not mandatory for scene text recognition systems to be ``end-to-end", because in some cases the detected image patches in the  bounding boxes predicted by scene text detectors may be too vague or tiny to be recognized. Yet the advantage of ``end-to-end"  approaches could be the internal feedback and the seamless interaction between the detection and recognition modules. 

\textbf{A. Scene Text Detection}

Most of the early DL-based approaches to scene text detection only support horizontal text detection. 

In~\cite{Seglink}, Shi et al. propose to detect texts with segments and links. They first detect a number of text parts, then predict the linking relationships between neighboring parts to form text bounding boxes. 

CTPN~\cite{CTPN} first detects text in sequences of fine-scale proposals, then recurrently connects these sequential proposals using BLSTM. 

TextBoxes~\cite{Textboxes} was designed based on SSD; but it adopts long default boxes that have large aspect ratios (as well as vertical offsets), because texts tend to have larger aspect ratios than general objects.  It only supports the detection of horizontal (vertical) texts in the beginning, later on,  the same authors propose TextBoxes++~\cite{TextBoxesplus} to support multi-oriented scene text detection. 

TextBoxes++ improves upon TextBoxes by replacing the rectangular box representation in conventional object detector by a quadrilateral representation. Moreover, the authors adopt a few long convolutional kernels to enable long but narrow receptive fields. It directly outputs word bounding boxes at multiple layers by jointly predicting text presence and coordinate offsets to anchor boxes. 

EAST~\cite{EAST} is a  U-shape fully convolutional network for detecting multi-oriented texts, it uses the PVANet to speed up the computation. 

%***deep direct regression***

\textbf{B. Scene Text Recognition}

In~\cite{CRNN}, Shi et al. propose to use CNN and RNN to model image features and  Connectionist Temporal Classification (CTC)~\cite{CTC} to transcript the feature sequences into texts. In~\cite{ShiWLYB16}, Shi et al. recognize scene text via attention based sequence-to-sequence model.

In~\cite{SlidingCNN}, Yin et al. propose the sliding convolutional character model, in which a sliding window is used to  transform a text-line image into sequential character-size crops. Then for each crop (of character-size, e.g. 32*40), they extract deep features using convolutional neural networks and make predictions.  These outputs from the sequential sliding windows are finally decoded with CTC.  Sliding CNN can avoid the gradient vanishing/exploding in training RNN-LSTM based models. 

\textbf{C. End-to-End Frameworks}

In~\cite{LiWS17,DeepText}, two end-to-end methods were proposed to localize and recognize text in a unified network, but they require relatively complex training procedures. 

In~\cite{MaskTextSpotter}, the authors design an end-to-end framework which is able to detect and recognize arbitrary-shape (horizontal, oriented, and curved) scene texts.

\textbf{Short Summary}. It is noticeable that, most state-of-the-art approaches for scene text detection and recognition focus on English language, but very little effort has been put into Chinese scene text recognition. 

\subsection{Related Datasets (English, Chinese) }

For English scene text detection, ICDAR2013, ICDAR2015~\cite{ICDAR2015}, COCO-Text~\cite{COCOText} are well-known real-world datasets, while SynthText~\cite{Synthetic} is a  commonly used synthetic English  scene text dataset. The training sets  of ICDAR2013 and ICDAR2015 are rather small, which are 229 and 1,000, respectively. COCO-Text has 43,686 training images (yet the annotations of some images are not very accurate), whereas SynthText has 800,000 images with 8 Million synthetic cropped  image patches.

For Chinese scene text detection and recognition, the three most related datasets to ours are RCTW~\cite{RCTW}, CTW~\cite{CTW} and ICPR 2018 MTWI challenge dataset~\cite{MTWI},  all of them were lately released (i.e., in 2017 and 2018). 

RCTW (a.k.a.CTW-12k) is an ICDAR-2017 competition dataset for scene text detection and recognition.  It has 12,263 annotated images, in which 8,034 images are used as training data. Most images in RCTW were captured using smart phone cameras; but it also includes screen-shot images from computers and smart phones, so these images are ``born-digital". The texts in the images of RCTW were annotated at the level of text lines using quadrilaterals to enable multi-oriented scene text detection.

CTW is a large dataset of Tecent street view images in China. It has 32,285 natural images with 1,018,402 Chinese characters, which is much larger than previous datasets including RCTW. All the natural images in CTW have the same resolution of 2048*2048 and all the street view  images were captured at fixed intervals (10 to 20 meters). Moreover, if two successive images have 70\% overlap, either of them was removed. For each Chinese character in a  natural scene image, they annotate its underlying character, its bounding box, and 6 attributes to indicate whether it is occluded, background complex or not, distorted or not, 3D raised or not, wordart or not, and handwritten or not, respectively.  The CTW images were annotated in crowd-sourcing manner by a third-party image annotation company, yet characters in English and other languages were not annotated. Finally, CTW has 3,850 unique Chinese characters (categories), 13.2\% of the character instances are occluded, 28.0\% have complex background, 26.0\% are distorted, and 26.9\% are 3D raised text. The authors argue that the majority of the 3,850 character categories are rarely-used and many of them have very few samples in the training data (imbalanceness),  so they only consider the recognition performance of the top 1000 frequent  characters in their benchmark experiments. 

\textbf{Discussion}.  DL-based methods are data-driven, they need to consume huge amount of data (images) to achieve good recognition performance. Therefore, a fundamental question for DL-based Chinese Photo OCR is: how many Chinese images do the DL-based methods need to achieve high recognition accuracy on the Chinese characters?  English language has only 26 letters, but it needs SynthText which has 800,000 synthetic images to achieve good English text recognition accuracy (around 80\%). While there are 6,763 commonly used Chinese characters, then the DL-based Chinese OCR methods need 200 Million images with Chinese texts to reach the same recognition accuracy as DL-based English photo OCR approaches? Due to the difficulty of collecting/annotating such huge number of photos, researchers need to investigate how to artificially generate photos with Chinese sentences.  Another simpler solution is to use tools such as~\cite{Synthetic} to generate the synthetic Chinese scene images. Indeed, different methods for generating synthetic scene images with Chinese texts/characters should be jointly adopted to enhance the diversity,  complexity and difficulty of the generated Chinese scene images.  Moreover, semi-supervised approaches can also be helpful, to fully utilize the limited annotated images and the vast amount of un-annotated ones, especially Web images from the Internet. 

\section{The ShopSign Dataset}
%diverse
%complex layout, difficult
\subsection{Dataset Collection and Annotation}
In developed countries such as USA, Italy and France, there are very limited number of characters in the shop signs, and the sizes of the shop signs are usually small. Moreover, the background of most shop signs in these countries are bare walls (whereas most Chinese shop signs use curtain/glass/wooden backgrounds). Owing to the differences in language, culture, history and the degree of development, the Chinese shop signs have distinctive features and high recognition difficulty. Even inside China, there is a big diversity in the materials and styles of the shop signs across different regions. For instance, in  major cities such as Shanghai and in the downtown of many cities, the shops usually adopt fiber-reinforced plastic (FRP) and neon sign boards; but in the suburb or developing regions, economic wooden and outdoor inkjet and acrylic shop signs are very common. The styles of the shop signs also vary in different provinces, e.g., shop signs in Inner Mongolia and the northwestern Xinjiang provinces are significantly different from Shanghai. 

\begin{figure*}
	\begin{center}
		\includegraphics[width=0.80\linewidth]{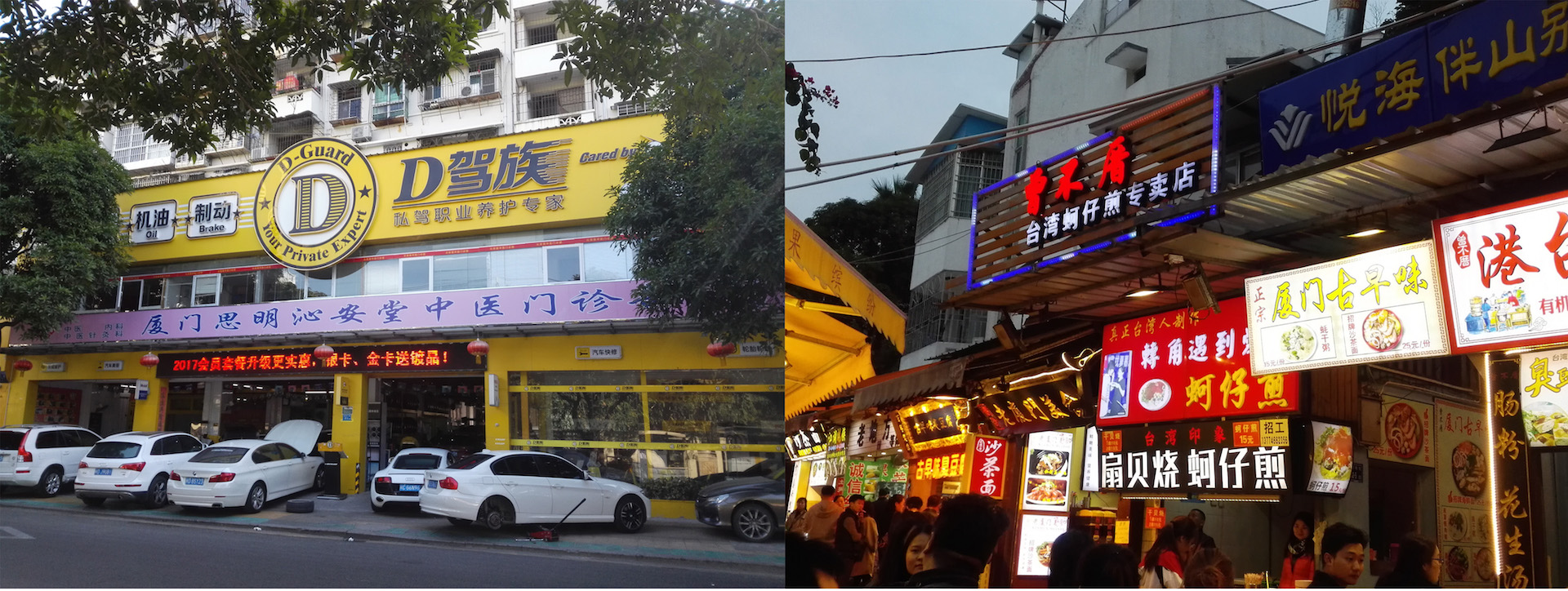}
		\includegraphics[width=0.80\linewidth]{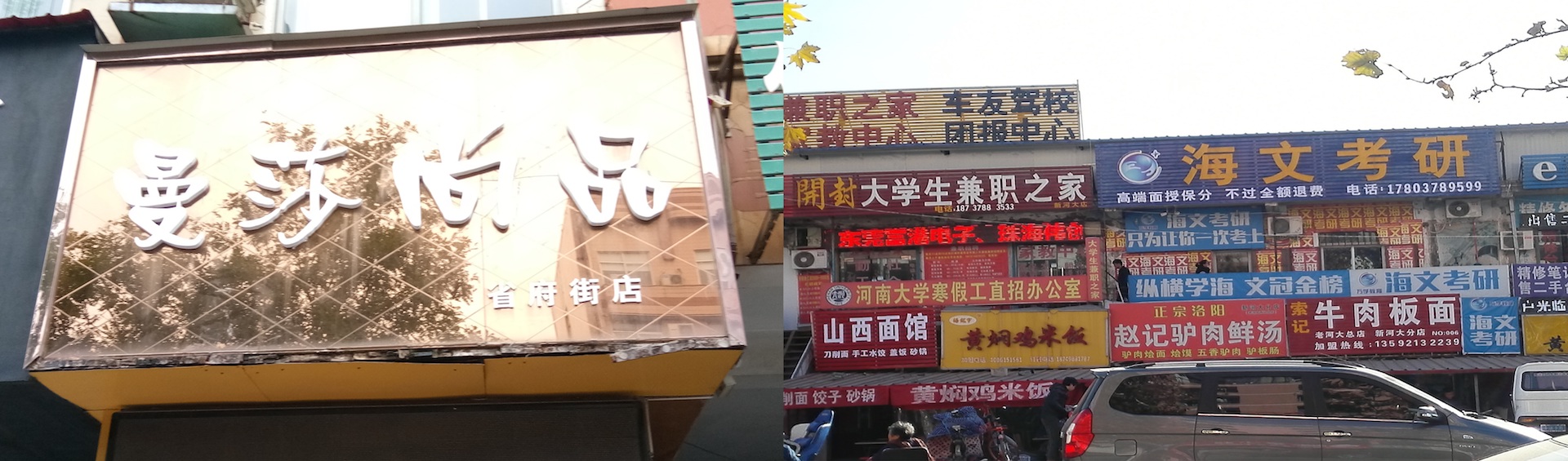}
		\includegraphics[width=0.80\linewidth]{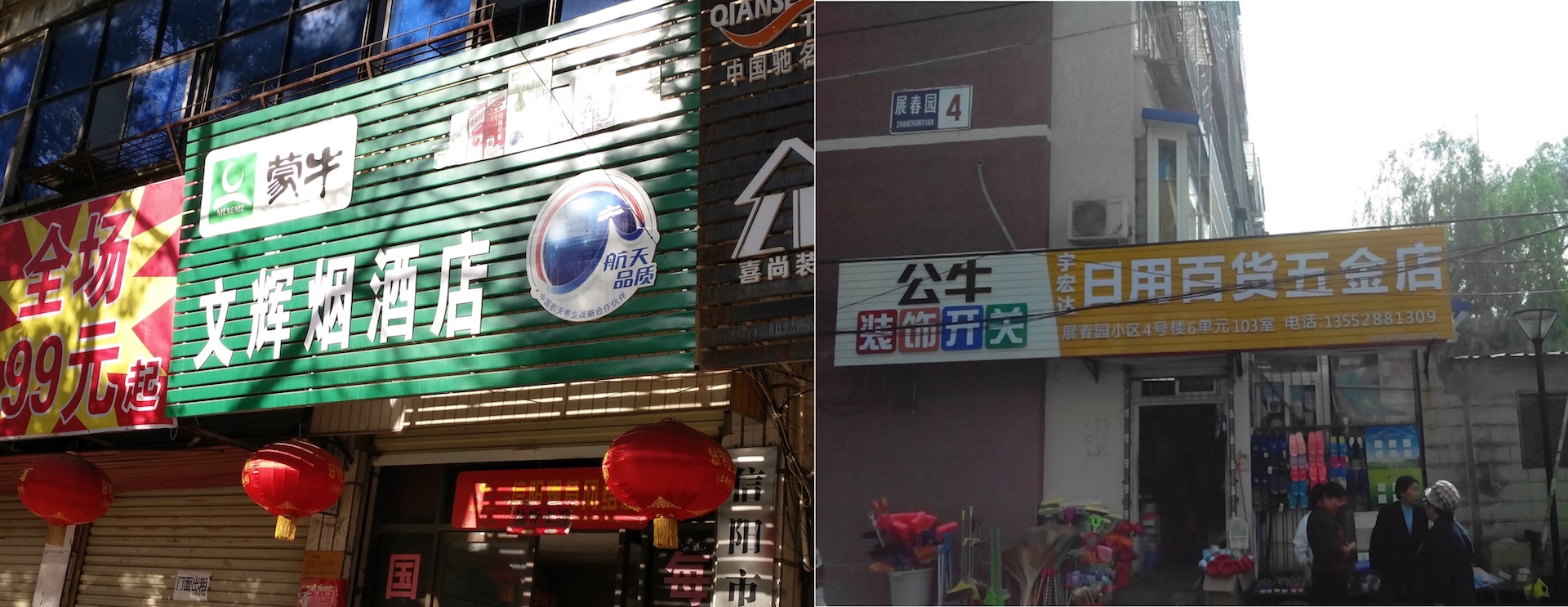}
	\end{center}
	\caption{Sample images of ShopSign. }
	\label{fig9}
\end{figure*}

\begin{figure*}
    \centering
    \begin{subfigure}[t]{0.19\textwidth}
        \includegraphics[width=\textwidth]{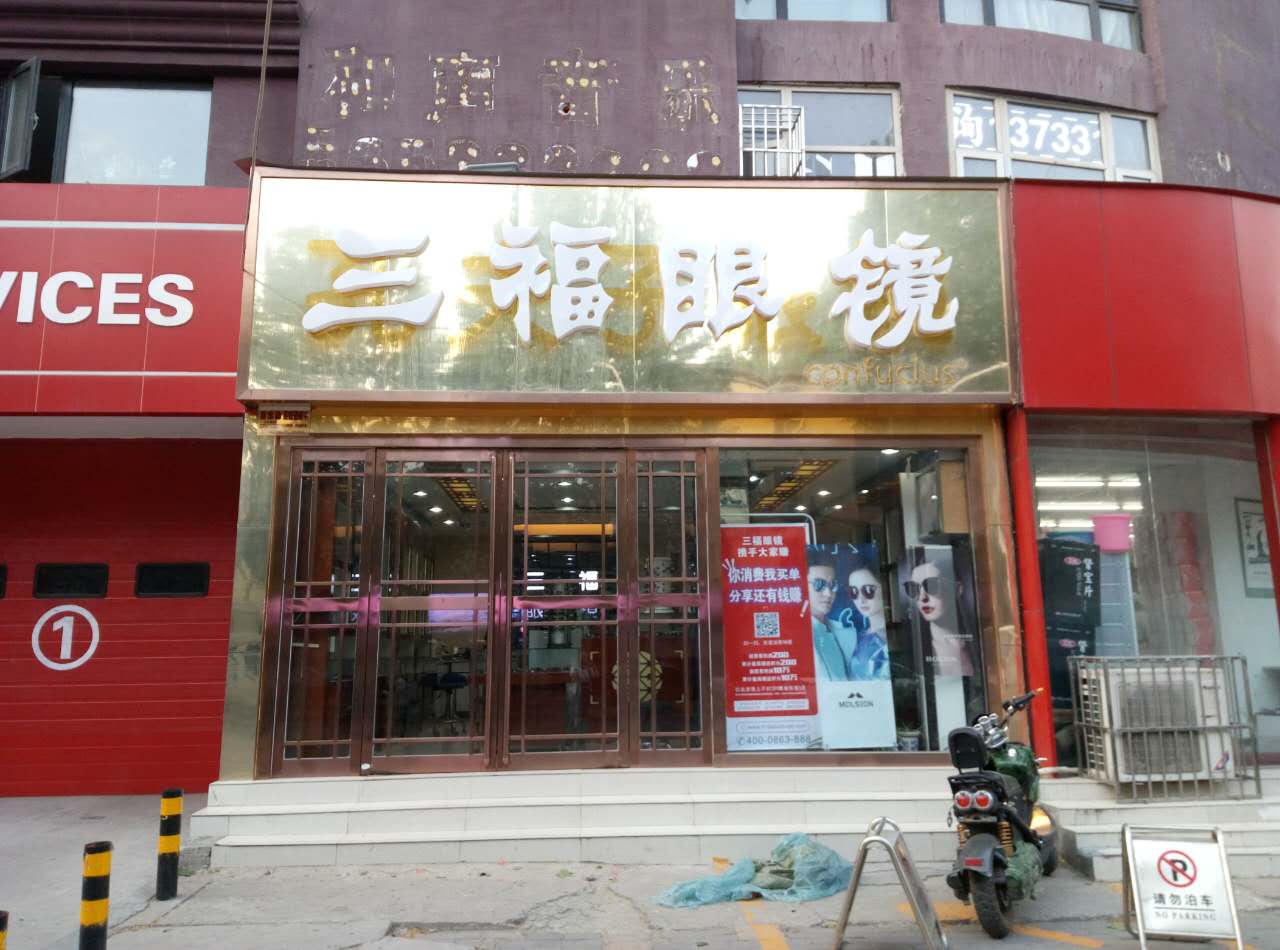}
        \caption{mirror}
        \label{fig:hard1}
    \end{subfigure}
    \hfill %add desired spacing between images, e. g. ~, \quad, \qquad, \hfill etc. 
    \begin{subfigure}[t]{0.19\textwidth}
        \includegraphics[width=\textwidth]{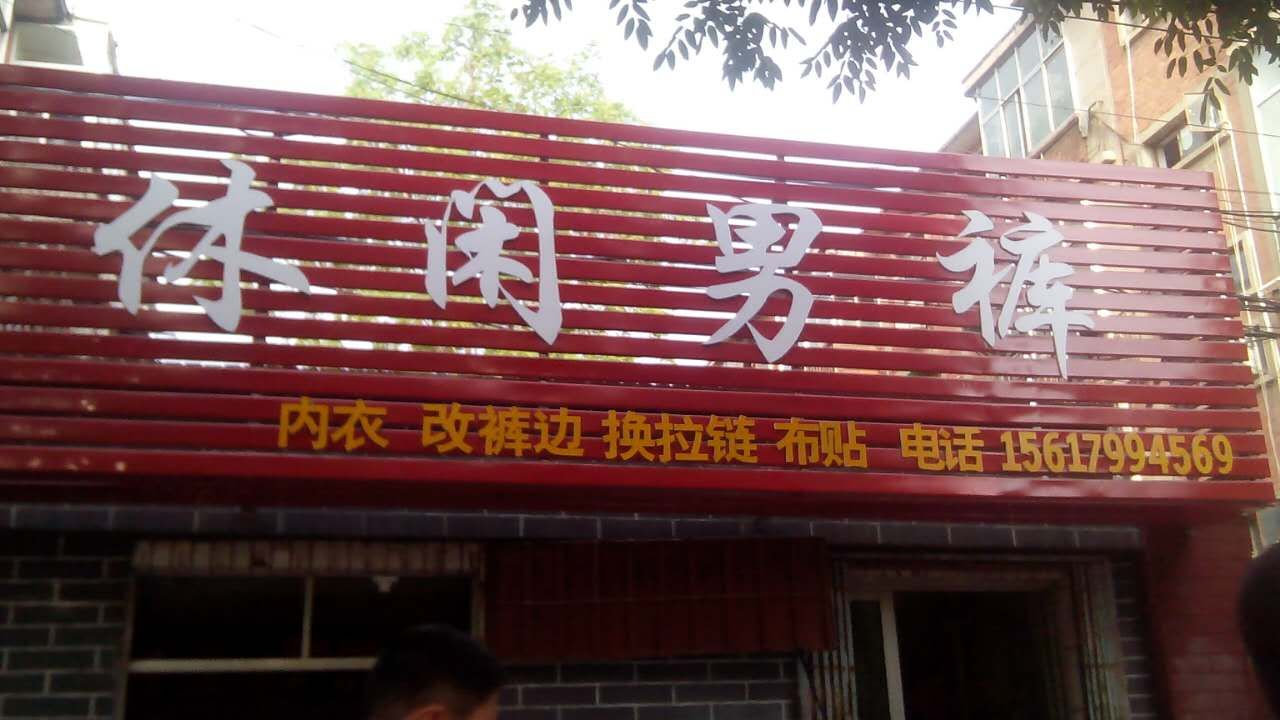}
        \caption{wooden}
        \label{fig:hard2}
    \end{subfigure}
    \hfill 
    \begin{subfigure}[t]{0.19\textwidth}
        \includegraphics[width=\textwidth]{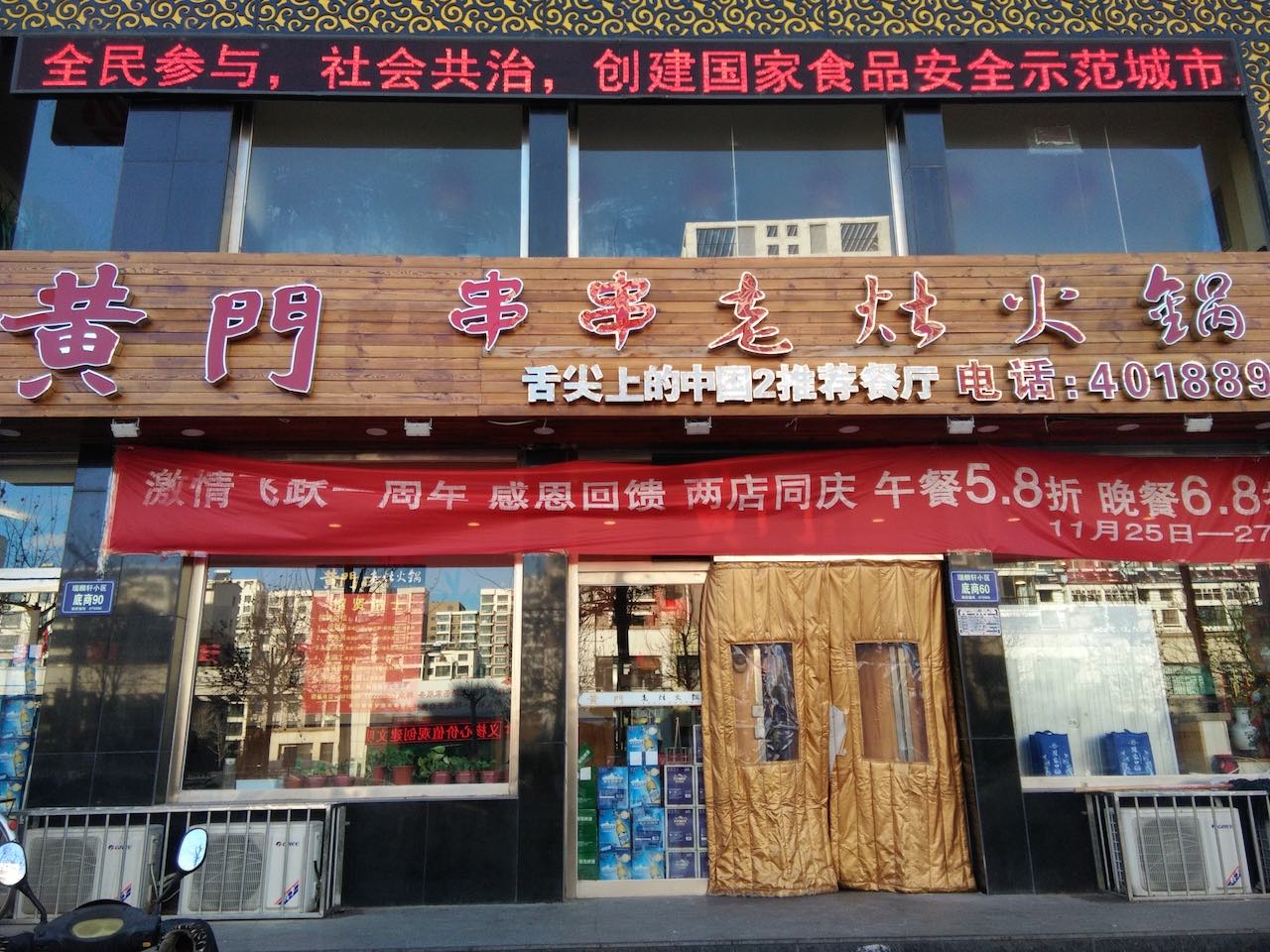}
        \caption{deformed}
        \label{fig:hard3}
    \end{subfigure}
    \hfill
    \begin{subfigure}[t]{0.19\textwidth}
        \includegraphics[width=\textwidth]{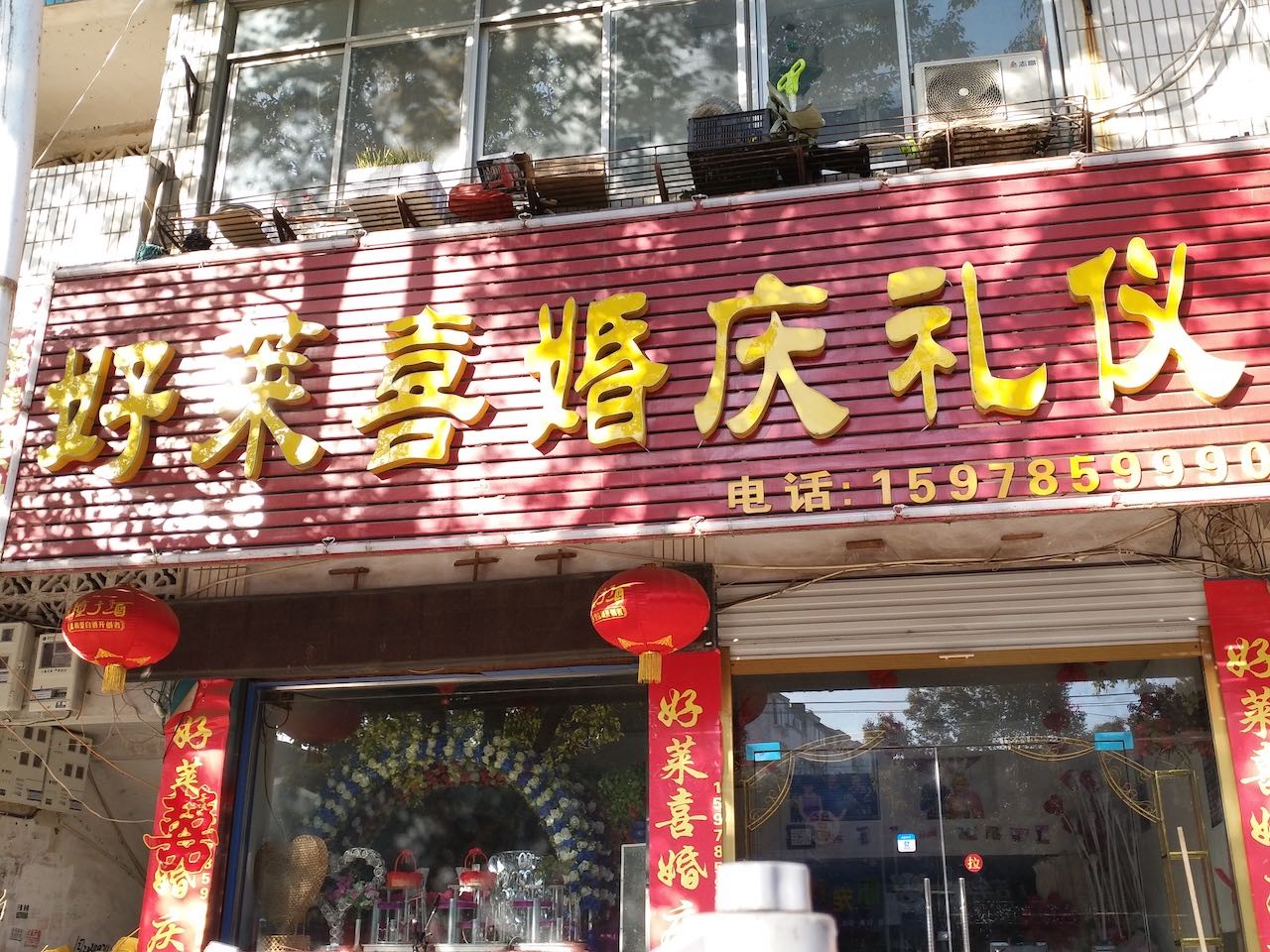}
        \caption{exposed}
        \label{fig:hard4}
    \end{subfigure}
    \hfill
    \begin{subfigure}[t]{0.19\textwidth}
        \includegraphics[width=\textwidth]{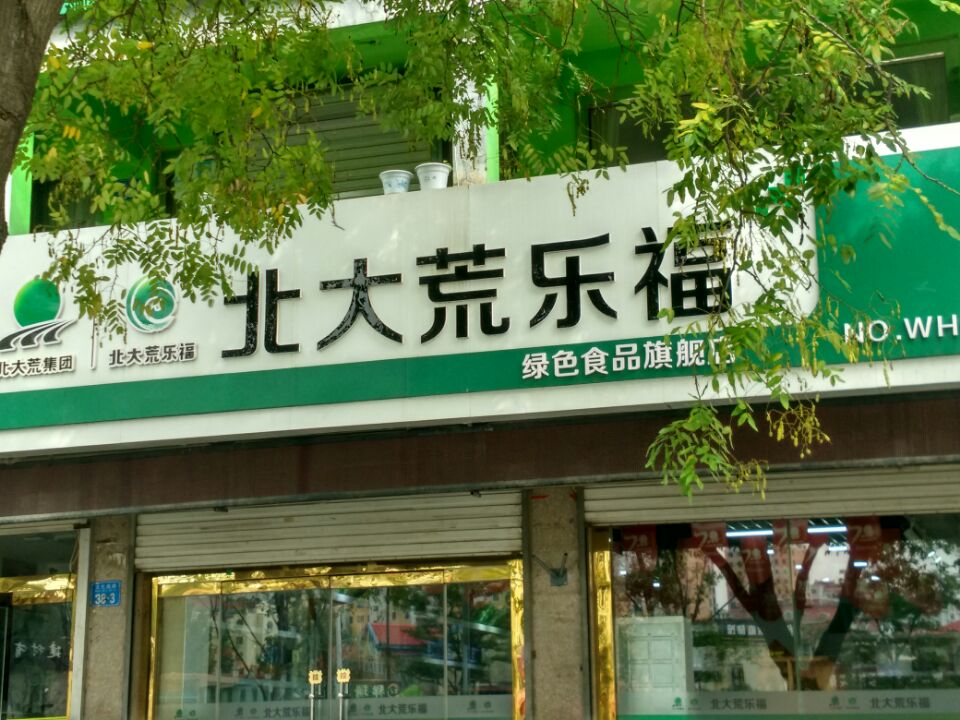}
        \caption{obscured}
        \label{fig:hard5}
    \end{subfigure}        
    \caption{Examples of the 5 categories of hard images.}
    \label{fig:hard}
\end{figure*}

Building a large-scale dataset of Chinese shop signs is a fundamental yet critical task that needs  enormous manual collection and annotation effort. We asked help from 40 students from our institution to collect shop sign images in various cities/regions of China, including Shanghai, Beijing, Inner Mongolia, Xinjiang, Heilongjiang, Liaoning, Fujian (Xiamen) as well as several cities/towns in Henan Province (Zhengzhou, Kaifeng, Xinyang, and a few counties/towns in Shangqiu and Zhoukou cities), with a duration of over two years. A total of 50 different cameras and smart phones were used in the collection and many of the images carry GPS locations. After the collection of the images, two faculty members and ten graduate students were involved in the annotation of these images (in text-line manner using quadrilaterals), which took another three months. Finally, the ShopSign dataset we build contains 25,770  well annotated images of Chinese shop sign in street views. In Figure \ref{fig9}, we showcase a few representative images in ShopSign.

\subsection{Dataset Characteristics}

\begin{table}[!ht]
	\begin{center}
		\begin{tabular}{|l|c|}
			\hline
			Item & Number
			/ Ratio
			\\
			\hline\hline
			\textbf{Total Number of Images} &  \textbf{25,770}\\
			Training images  & 20,738 \\
			Testing images  & 5,032 \\
			Text-lines & \textbf{196,010} \\
			Text-lines in Training set & 146,570\\
			Text-lines in Testing set & 49,440\\
			Chinese Characters & \textbf{626,280}\\
			Unique Chinese Characters & \textbf{4,072} \\
			\hline\hline
			\textbf{Number of Unique Characters with} & \\
			1          occurrence & 537 \\
			2-10     occurrences & 1,150 \\
			11-50   occurrences & 958 \\
			51-100   occurrences & 395 \\
			101-200 occurrences & 338 \\
			201-500 occurrences & 360 \\
			501-1000 occurrences & 188 \\
			1001-2000 occurrences & 108 \\
			2001-3000 occurrences & 22 \\
			3001-5000 occurrences & 15 \\
			$ \geq 50$    occurrences  & 1,439 \\
			$ \geq 100$    occurrences  & 1,039 \\
			$ \geq 200$    occurrences  & 694 \\
			$ \geq 500$    occurrences  & 333 \\
			$ \geq 1000$    occurrences  & 145 \\
			$ \geq 2000$    occurrences  & 37 \\
			\hline\hline
			\textbf{Ratio of Unique Characters with} & \\
			1           occurrence & 13.2\% \\
			2-10     occurrences & 28.2\% \\
			11-50   occurrences & 23.5\% \\
			$ \leq 50$    occurrences  & 65\% \\
			$ \leq 100$    occurrences  & 74.7\% \\
			101-500 occurrences & 17.1\%\\
			$ \geq 500$    occurrences  & 8.2\% \\
			$ \geq 1000$    occurrences  & 3.6\% \\
			$ \geq 2000$    occurrences  & 0.9\% \\
			\hline\hline
			\textbf{Number of Occurrences  for} & \\
			No. 1  most frequent character & 7,603 \\
			No. 2 most frequent characters & 6,503 \\
			No. 3 most frequent characters & 5,276 \\
			No. 4 most frequent characters & 5,121 \\
			No. 5 most frequent characters & 5,074 \\
			\hline\hline
			\textbf{Distribution (Class) Imbalance} & \\
			Characters with $ \geq 500$   occurrences  & 8.2\%\\
			    Their Total number of occurrences  &64.7\%\\
			Characters with $ \leq 100$   occurrences  & 74.7\%\\
			    Their Total number of occurrences  &9.5\%\\
			\hline
		\end{tabular}
	\end{center}
	\caption{Basic Statistics of ShopSign.}
	\label{tab8}
\end{table}

In Table \ref{tab8}, we present the basic statistics of ShopSign.  It contains 25,770 Chinese natural scene images and 196,010 text-lines.   The total number of unique Chinese characters is 4,072, with 626,280 occurrences in total. 
Overall, ShopSign has the following characteristics.

\begin{enumerate}
\item \textbf{Large-scale}. It contains both images with horizontal and  multi-directional texts. It comprises more than 10,000 images with horizontal texts and over 10,000 images having multi-directional texts.  

\item \textbf{Night images}. It includes near 4,000 night images (captured in the night). In these night images, the sign boards are very remarkable and the rest background areas are comparatively dark. Such night images rarely exist in other datasets. 

\item \textbf{Special categories of hard images}. It consists of 5 special categories of hard images, which are mirror, wooden, deformed, exposed and obscure, as depicted in Figure \ref{fig:hard}. Text detection and recognition over such hard images should be more challenging than ordinary natural scene images.  Besides, it has 500 very difficult scene images, with complex layout or gloomy background, as we can observe from Figure \ref{fig9}.

\item \textbf{Sparsity and class imbalance}. The number of unique characters with 500 or more occurrences is 333 (8.2\% in ratio), but the sum occurrences  of these characters occupies 64.7\% of the total occurences of the characters in the dataset. In comparison, the ratio of characters with 100 or less occurrences is 74.7\%, but their total number of occurrences is only 9.5\% in the dataset. Furthermore, 537 characters only have 1 occurrence, and 1,687 characters (41.4\% in ratio) have 10 or less occurrences. Hence, the distribution of the Chinese characters' occurence frequency in ShopSign is highly skewed. 

\item \textbf{Diversity}. The dataset spans several provinces in China, from Beijing and Shanghai to northeastern and northwestern provinces in China, from downtown to developing regions. The photos were captured in different seasons of the year, using 50 different smart phones. The styles of the sign boards, their backgrounds and texts are also very diverse.  

\item \textbf{Pair images}. Our dataset contains 2,516 pairs of images. In each pair of images, the same sign board was shot twice, from both frontal and tilted perspectives. Pair images facilitate the evaluation/comparison of an algorithm's performance on horizontal and multi-oriented text detection.
\end{enumerate}

In Table \ref{tab1}, we make comparisons between our ShopSign dataset and the existing CTW dataset, which is the most relevant dataset to ours.
  
%introduce something about the RCTW/MTWI dataset characteristics

\begin{table*}[ht]
	\begin{center}
		\begin{tabular}{p{3cm}|p{5.5cm}|p{7.5cm}}
			\hline
			& CTW & ShopSign \\
			\hline\hline\hline
			Themes & Street Views (roads, buildings, trees, etc.) & Shop Signs (sign boards) with texts \\
			\hline
			Equipments & StreetView Collection Vehicles with Identical Nikon SLR Cameras & Smart Phones. 50 different smart phones from various brands (e.g. iPhone, Huawei, Samsung, Vivo, Xiaomi, etc.) \\
			\hline
			Collection Manner & Automated (by vehicles) & Manually (by 40 different research assistants) \\
			\hline
			Shooting Angles & fixed angles &  arbitrary angles \\
			\hline
			Capture Distance & fixed distance: $\geq$ 10 meters (vehicles inside motor vehicle lanes)  & Varying distances (to the targets), 2-8 meters, on pavements \\
			\hline
			Time Span & around 3 months & 2 years and 4 months \\
			\hline
			Sites & a few major cities (developed) & Wide geographical coverage (Beijing, Shanghai, Xiamen, Xinjiang, Mongolia, Mudanjiang, Huludao and a few cities and small towns in Henan province) including many developing regions or small towns where street view vehicles do not reach. \\
			\hline			
			Photo Resolutions & unanimously 2048*2048 &  Various resolutions (3024*4032, 1920*1080, 1280*720, ...) \\
			\hline
			Annotation Methods & Third-party crowd-sourcing platform, per character. Suffer from well-known quality issues caused by crow-sourcing. & 10 full-time research assistants (manually annotated, and with calibrations), per textline.  Highly precise annotations. \\
			\hline
			Diversities of the Sign Boards & Limited (Medium or Expensive Sign Boards) & Various materials for sign boards, including many inexpensive (e.g., cloth or wood) ones used by rural or developing urban areas. \\
			\hline
			Languages & Chinese only  & Chinese and English \\
			\hline
			Richness of texts &  Medium or large signboards, no tiny ads; no indoor texts; no tiny texts; no texts in the mirror boards. & Various scales of text areas: embossed texts on the buildings; texts on the mirrors; indoor texts; tiny ads; rich texts on sign boards of different styles. \\
			\hline
			Night Photos & No & Yes, with near 4000 photos shoot in the night. \\
			\hline
			Pair Photos & No & Yes, with 2,516 pairs of photos, in which two photos were taken for one shop sign from  horizontal and multi-oriented angles. \\
			\hline
			Very difficult photos & very few & More than 500 very difficult photos. \\
			\hline
		\end{tabular}
	\end{center}
	\caption{Comparison Between CTW and our ShopSign dataset.}
	\label{tab1}
\end{table*}

\subsection{Dataset Split}
The ShopSign dataset will be completely shared with the research community. We will not reserve the testing set but release it to the public.  However, we suggest researchers to train and adjust parameters on the training set, and minimize the number of runs on the testing set. 

%Since detect detection and recognition have different dataset requirements and evaluation criteria, we separately split the dataset for benchmarking purposes. 

First,  we split ShopSign into training (Train1) and testing set (Test1), which contain 20,738 and 5,032 images, respectively. The testing set comprises 2,516 pairs of images, which were independently collected/annotated in a later period (by a different group of research assistants) than the training set. We collected the testing set in such a way that the real performance of state-of-the-art text detection algorithms  is more authentically reflected. The pair images can also compare and reveal the ability of an algorithm in detecting horizontal and multi-oriented scene texts.  Train1 and Test1 will also be used for assessing the performance of text recognition algorithms. The collection of cropped text-lines from Train1 (since they have annotations) will be used for training text recognition models, while the set of cropped text-lines from Test1 will be used for testing their recognition performance.  It should be noted that, due to the large number of classes in ShopSign (with 4,072 unique characters) and the text-line based annotation manner, the data is very sparse and imbalanced (the class imbalance issue in the number of samples for different Chinese characters has been disccused above).  

Second, for specific evaluation of text detection performance on the five ``hard" categories of images, ShopSign is re-split into another training (Train2) and testing set (Test2). Test2 comprises half of the images from each of the five ``hard" categories, whereas all the other images of ShopSign are used as the new training set, i.e., Train2. 

In short, Train1 and Test1 are for evaluating both the text detection and recognition algorithms; whereas Train2 and Test2 are specially designed for assessing the performance of text detection algorithms on the five ``hard" categories of images. 

In view of the large-scale Chinese characters (i.e. 4,072 classes in our dataset), our ShopSign dataset is very sparse and imbalanced (as depicted in the above subsection). What is more, the text-line based annotation manner makes the data even more sparse and challenging.

\section{Experiments}

Having the ShopSign dataset at hand, we now proceed to illustrate how it helps improve Chinese scene text detection and recognition results. 

%As we will also see, ShopSign is particularly effective at improving text detection results on mirror text-lines, exposed ones,  obscured ones, wooden ones and deformed ones. 

\subsection{Influence of language on text dection}
For scene text detection,  people only need to localize the areas of the candidate text lines (or areas), without recognizing the  content (characters) of the texts. Furthermore, some text lines may be too tiny or vague so that they can only be detected as textual lines/areas but the texts (characters) are too unclear to be recognized. Hence, the intrinsic features of the images utilized by deep learning based scene text detection algorithms should be very different from those used by deep learning based text recognition algorithms. 

Early research on scene text detection only focuses on English natural scene texts.  But many researchers in this field often  raise the following questions: do text detection models, trained on English natural scene images, perform well on the Chinese ones and vice versa? Does the performance of scene text detection algorithms differ between English natural scene images and the Chinese ones? 

In order to investigate whether text language has a significant influence on the performance of scene text detection algorithms, in Table \ref{tab2}, we first present the text detection results of the EAST and TextBoxes++ models trained on SynthText,  on the test set of ShopSign (Test1). We see that these two models  achieve very low recall and precision results on ShopSign; their recall rate are only 12\%-15\%.  This demonstrates that the models trained by deep learning based scene text detection algorithms on the English natural scene images perform very poorly on the Chinese ones. Therefore, text language has a very significant influence on the performance of scene text detection algorithms. 

Thus,  for Chinese scene text detection, datasets with Chinese natural scene images are needed. 

\subsection{Baseline Experiments on ShopSign}
We next report the performance of baseline text detection algorithms on ShopSign, using Train1 and Test1. In Table \ref{tab2}, we  report the prediction performance of the major text detection algorithms (using Train1 of ShopSign) on the test set (i.e., Test1) of ShopSign, which contains 2,516 pairs of images, with each pair of images containing both horizontal and multi-oriented images captured for the same shop sign(s).  The pair images can more comprehensively evaluate the relative performance of scene text detection models, since the two images were shot for the same shop sign but from different perspectives (one frontal, the other leaning).  We observe that, EAST achieves the best text detection performance  on ShopSign, with a recall of 57.9\%-58.4\%. We also notice that CTPN obtains better text detection results than TextBoxes++ on the horizontal test set of ShopSign, but the latter outperforms the former in multi-oriented scene texts of ShopSign. Therefore, CTPN is more capable in horizontal text detection. 

\subsection{Cross-dataset Text Detection Performance}

We now compare the relative difficulty of ShopSign with RCTW/CTW/MTWI on scene text detection, using EAST, TextBoxes++ and CTPN. We choose the RCTW/CTW/MTWI datasets to do cross-dataset generalization because they are the only available scene text datasets that contain Chinese texts/characters. 

%as the baseline algorithms owing to their state-of-the-art performance on scene text detection. 

Cross-dataset generalization is an evaluation for the generalization ability of a dataset. Because there are no officially released ground-truth (labels) for the test datasets of  RCTW/CTW/MTWI, we use their official training  data to train the corresponding text detection models, then test these models on test set (i.e., Test1) of ShopSign.

In Table \ref{tab2}, we use three scene text detection algorithms to train models on RCTW/CTW/MTWI, then report their  performance on the test set of ShopSign.  We see that, text detection models trained on CTW and MTWI by all the three algorithms only obtain a recall of 16.1\%-38\% on ShopSign, which is very low. This indicates that, ShopSign is a more challenging Chinese natural scene image set than CTW and MTWI.	We also see that, EAST trained on RCTW obtains the best recall result on ShopSign, which is  between 50.5\% and 53.2\%, while the other two algorithms only have a recall below 44.2\%.

In short, ShopSign is a more comprehensive and difficult Chinese natural scene text dataset than RCTW/CTW/MTWI. 

%recall, precision, h-mean
%recall, precision, h-mean
\begin{table*}[ht]
	\begin{center}
		%\begin{tabular}{l|l|l}
		\begin{tabular}{|c|l|l|l|l|l|l|l|}
			\hline
			 \multirow{2}{*}{Datasets} &  \multirow{2}{*}{Methods}& \multicolumn{3}{c|}{Horizontal} & \multicolumn{3}{c|}{Multi\_Oriented}\\
			 \cline{3-8} 
			 &&R &P &H &R &P &H\\
			\hline\hline\hline
			\multirow{2}{*}{SynthText}
			&EAST  &0.124& 0.125& 0.124   &  0.133 & 0.174 & 0.150\\
				 \cline{2-8} 
			&TextBoxes++  & 0.115 & 0.287 & 0.165&0.150 & 0.330 & 0.206\\
			\hline\hline
			
			  \multirow{3}{*}{ShopSign}       	
			&EAST  &0.584 &0.364 &0.448 & 0.579 &0.410 &0.480\\
			\cline{2-8} 
			&TextBoxes++  &0.471 &0.501 &0.486 &0.479 &0.476 &0.478\\
			\cline{2-8} 
			&CTPN &0.535 &0.566 &0.550 &0.444 &0.518 &0.478\\
			\hline\hline
			
			\multirow{3}{*}{CTW}
			&EAST  &0.241 &0.261 &0.250 &0.215 &0.279 &0.243\\
				 \cline{2-8} 
			&TextBoxes++ &0.346 &0.084 &0.135 &0.313 &0.075 &0.121\\
			 \cline{2-8} 
			& CTPN  &0.182 &0.371 &0.244 &0.161 &0.379 &0.226\\
			\hline\hline
			
			\multirow{3}{*}{RCTW}	
			&EAST  &0.532 &0.371 &0.437 &0.505 &0.412 &0.454\\
				 \cline{2-8} 
			&TextBoxes++  &0.401 &0.516 &0.451 &0.380 &0.432 &0.405\\
          	 \cline{2-8} 
			&CTPN &0.442 &0.446 &0.444 &0.373 &0.407 &0.389\\
            \hline\hline
            
             \multirow{3}{*}{MTWI}        	
			&EAST  &0.360 &0.250 &0.295 &0.319 &0.274 &0.294\\
				 \cline{2-8} 
			&TextBoxes++ &0.328 &0.392 &0.357 &0.306 &0.34 &0.322\\
		   	 \cline{2-8} 		
			&CTPN &0.380 &0.490 &0.428 &0.336 &0.480 &0.395\\
        	\hline\hline
   
		\end{tabular}
	\end{center}
\caption{ Results of State-of-the-art Text Detection Methods on the test dataset of ShopSign.}
\label{tab2}
\end{table*}

\subsection{Specific Performance on the Hard Categories}

%recall, precision, h-mean
\begin{table*}[ht]
	\begin{center}
		%\begin{tabular}{l|l|l}
		\begin{tabular}{|c|l|l|l|l|l|l|}
			\hline
			Methods & Datasets&Mirror &Wooden &Deformed &Exposed &Obscured\\
			\hline\hline
			\multirow{6}{*}{EAST}
			& CTW &0.155	&0.242	&0.219	&0.274	&0.209\\
			\cline{2-7}	
			& CTW+ShopSign	&0.515	&0.561	&0.561	&0.534	&0.549\\	
			\cline{2-7}	
			& MTWI	&0.295	&0.328	&0.351	&0.347	&0.329\\	
			\cline{2-7}	
			&MTWI+ShopSign	&0.542	&0.542	&0.560	&0.574	&0.542\\
			\cline{2-7}	
			& RCTW	&0.452	&0.467	&0.504	&0.490	&0.485\\	
			\cline{2-7}	
			& RCTW+ShopSign	&0.533	&0.558	&0.569	&0.595	&0.558\\	
			\hline \hline 
			
			\multirow{6}{*}{TextBoxes++}
			&CTW &0.324 &0.287 &0.285 &0.300 &0.318\\
			\cline{2-7}	
			& CTW+ShopSign	&0.494	&0.398	&0.391	&0.440	&0.422	\\
			\cline{2-7}	
			& MTWI	&0.360	&0.272	&0.273	&0.297	&0.291	\\
			\cline{2-7}	
			& MTWI+ShopSign&	0.527	&0.404	&0.394	&0.466	&0.431\\
			\cline{2-7}		
			& RCTW	&0.449	&0.334	&0.330	&0.397	&0.357	\\
			\cline{2-7}	
			&RCTW+ShopSign &0.515	&0.397	&0.387	&0.478	&0.431\\	
			\hline \hline
			
			\multirow{6}{*}{CTPN}	
			&CTW	&0.136	&0.152	&0.167	&0.114	&0.173\\ 
			\cline{2-7}		
			& CTW+ShopSign	&0.565	&0.496	&0.465	&0.503	&0.507\\
			\cline{2-7}	
			& MTWI	&0.283	&0.333	&0.352	&0.334	&0.353\\ 
			\cline{2-7}	
			& MTWI+ShopSign	&0.564	&0.516	&0.486	&0.527	&0.511\\ 
			\cline{2-7}	
			& RCTW	&0.398	&0.371	&0.380	&0.404	&0.403\\ 
			\cline{2-7}	
			& RCTW+ShopSign	&0.565	&0.499	&0.489	&0.568	&0.522\\
			\hline \hline
		\end{tabular}
	\end{center}
	\caption{ Recall of Baseline Text Detection Methods on the difficult images of ShopSign (whole image level).}
	\label{tab3}
\end{table*}

%recall, precision, h-mean
\begin{table*}[ht]
	\begin{center}
		%\begin{tabular}{l|l|l}
		\begin{tabular}{|c|l|l|l|l|l|l|}
			\hline
			Methods & Datasets&Mirror &Wooden &Deformed &Exposed &Obscured\\
			\hline\hline
			\multirow{6}{*}{EAST}
			& CTW &0.096	&0.152	&0.201	&0.239	&0.112\\
			\cline{2-7}	
			& CTW+ShopSign	&0.376	&0.488	&0.462	&0.543	&0.341\\	
			\cline{2-7}	
			& MTWI	&0.196	&0.264	&0.316	&0.351	&0.155\\	
			\cline{2-7}	
			&MTWI+ShopSign	&0.388	&0.492	&0.496	&0.564	&0.343\\
			\cline{2-7}	
			& RCTW	&0.296	&0.389	&0.444	&0.452	&0.278\\	
			\cline{2-7}	
			& RCTW+ShopSign	&0.380	&0.492	&0.479	&0.585	&0.359\\	
			\hline \hline 
			
			\multirow{6}{*}{TextBoxes++}
			&CTW&0.244&0.373&0.333&0.282&0.302\\
			\cline{2-7}	
			& CTW+ShopSign	&0.488	&0.635	&0.427	&0.521	&0.473	\\
			\cline{2-7}	
			& MTWI	&0.356	&0.450	&0.359	&0.372	&0.320	\\
			\cline{2-7}	
			& MTWI+ShopSign&	0.524	&0.645	&0.466	&0.548	&0.486\\
			\cline{2-7}		
			& RCTW	&0.436	&0.535	&0.380	&0.468	&0.402	\\
			\cline{2-7}	
			&RCTW+ShopSign &0.500	&0.637	&0.432	&0.569	&0.482\\	
			\hline \hline
			
			\multirow{6}{*}{CTPN}	
			&CTW	&0.096	&0.177	&0.248	&0.133	&0.124\\ 
			\cline{2-7}		
			& CTW+ShopSign	&0.332	&0.373	&0.350	&0.362	&0.342	\\
			\cline{2-7}	
			& MTWI	&0.340	&0.401	&0.346	&0.356	&0.328		\\ 
			\cline{2-7}	
			& MTWI+ShopSign	&0.412	&0.455	&0.368	&0.415	&0.372	\\ 
			\cline{2-7}	
			& RCTW	&0.368	&0.420	&0.359	&0.426	&0.366		\\ 
			\cline{2-7}	
			& RCTW+ShopSign	&0.372	&0.420	&0.372	&0.367	&0.407	\\
			\hline \hline
		\end{tabular}
	\end{center}
	\caption{ Recall of Baseline Text Detection Methods on the difficult images of ShopSign (specific text-line level).}
	\label{tab4}
\end{table*}

To characterize the detection difficulty on the  five ``hard" categories of ShopSign, which are mirror, wooden, deformed,  exposed and obscured, we report the specific text detection results on each category, using Train2 and Test2. In Table \ref{tab3}, we first show the overall results of the three methods on each specific ``hard" category of ShopSign.  Without ShopSign, the three text detection algorithms again achieve low recall results on each`` hard" category, but the performance difference from what we observe from Table \ref{tab2} is not significant.  This is because the results from Table \ref{tab3} are image-level, they include all the text lines of the whole image that contains hard text lines (such as the  text lines with mirrors).  When ShopSign is combined with RCTW/CTW/MTWI, we observe very significant performance improvement.  Therefore, ShopSign dataset improves the performance of the scene text detection algorithms on the ``hard" examples. 

%, using the same models trained on RCTW/CTW/MTWI
%We see that EAST outperforms the other two detectors by a large margin, ***  Moreover, ***

To further check the performance of scene detection algorithms on the specific ``hard" examples of ShopSign images, in Table \ref{tab4} we pick the corresponding ``hard" text lines from each ``hard" image, and separately calculate their recall.  That is, for each image that belongs to the ``hard" category, we only measure the recall results of the specific hard text lines of the image. 

From Table \ref{tab4}, it is clear that the recall results of the scene text detection algorithms on the hard text lines of ShopSign is much lower than those reported in Table \ref{tab2}. This shows that, the five  specific ``hard" examples of ShopSign are more challenging. Moreover, we see that TextBoxes++ is less influenced by the hard examples than EAST and CTPN.  In particular, both EAST and CTPN perform poorly on the mirror and obscured images, with a recall under 41.2\%, while the performance of CTPN also drops on the exposed scene text images.

Overall, the ShopSign dataset is useful in the detection of ``hard" Chinese scene text images.

\section{Conclusion and Future work}
In this work, we present the ShopSign dataset, which is a diverse large-scale natural scene images with Chinese texts. 
We describe the properties of the dataset and carry out baseline experiments to show the challenges of this dataset. 
In future work, we will conduct experiments and report the performance of baseline text recognition algorithms on ShopSign.
We will also show the corresponding difficulties of text recognition on the ``hard" images of ShopSign.

Given the large number of Chinese characters, data sparsity is very common (even inevitable) in datasets of Chinese natural scene images, although the sizes of CTW and ShopSign are already very large. The text-line based annotation manner further increases  the data sparsity,  when we consider the combinations of Chinese  characters in a sequence. Besides, the number of samples for different characters is also highly imbalanced. A large-scale synthetic dataset of scene images with Chinese texts is needed by the community, which may contain tens of millions of images. To this end, we are designing GAN (generative adversarial network) based techniques to generate such synthetic datasets.

\section{Acknowledgments}
We are very grateful to the 40 students who contributed to the collection of the ShopSign images, including (but not limited to)  Mr. Yikun Han, Mr. Jiashuai Zhang, Mr. Kai Wu, Ms. Xiaoyi Chen, Mr. Cheng Zhang, Ms. Shi Wang, Ms. Mengjing Sun, Ms. Jia Shi, Ms. Xin Wang, Ms.  Huihui Wang, Ms. Lumin Wang, Mr. Weizhen Chen, Mr. Menglei Jiao, Mr. Muye Zhang, Mr. Zhiqiang Guo and Dr. Wei Jiang from NCWU, etc. We deeply acknowledge the great efforts in the annotation of the ShopSign dataset, made by Ms. Guowen Peng, Ms. Lumin Wang, Mr. Yikun Han, Mr. Yuefeng Tao, Mr. Jingzhe Yan, Mr. Hongsong Yan, Ms. Feifei Fu, Ms. Mingjue  Niu, etc. Ms. Guowen Peng has also spent a huge amount of time in the re-arrangement of the images, the merging and correction of the annotation results. We also thank the useful feedbacks from Prof. Xucheng Yin (USTB), Mr. Chang Liu, and Dr. Chun Yang.
%Without their help, it will be impossible to build this large-scale image set of Chinese natural scene texts. 
%-------------------------------------------------------------------------

{\small
\bibliographystyle{ieee}
\bibliography{shopsign}
}

\end{document}